\documentclass[]{elsarticle} 

\usepackage{lineno,hyperref}
\modulolinenumbers[5]

\usepackage{amsmath}
\usepackage{latexsym}

\begin{document}

\begin{frontmatter}

\title{Avoiding Confusion between Predictors and Inhibitors in Value Function Approximation}

\author[mymainaddress]{Patrick Connor \corref{mycorrespondingauthor}}
\ead{patrick.connor@dal.ca}
\author[mymainaddress]{Thomas Trappenberg}

\address[mymainaddress]{Faculty of Computer Science, Dalhousie University, 6050 University Avenue, \\ PO BOX 15000, Halifax, Nova Scotia, Canada, B3H 4R2}
\cortext[mycorrespondingauthor]{Corresponding author}
\begin{abstract}
In reinforcement learning, the goal is to seek rewards and avoid punishments.  A simple scalar captures the value of a state or of taking an action, where expected future rewards increase and punishments decrease this quantity.  Naturally an agent should learn to predict this quantity to take beneficial actions, and many value function approximators exist for this purpose.  In the present work, however, we show how value function approximators can cause confusion between predictors of an outcome of one valence (e.g., a signal of reward) and the inhibitor of the opposite valence (e.g., a signal canceling expectation of punishment).  We show this to be a problem for both linear and non-linear value function approximators, especially when the amount of data (or experience) is limited.  We propose and evaluate a simple resolution: to instead predict reward and punishment values separately, and rectify and add them to get the value needed for decision making.  We evaluate several function approximators in this slightly different value function approximation architecture and show that this approach is able to circumvent the confusion and thereby achieve lower value-prediction errors.
\end{abstract}


\end{frontmatter}

%




\section{Introduction}
Our world is full of sensory stimuli.  For reinforcement learning (RL), each configuration of such stimuli must be interpreted as expressing a particular world state.  In simple RL problems, it is possible to enumerate all such states and employ a table to store the state values.  It would seem that for most real world problems, the number of stimulus dimensions and the continuous quality of sensory information begs rather for the use of a function approximation approach to map a value onto each state (or state-action combination).  Value-function approximators \cite{SuttonBarto_1998} (VFAs) exactly fulfill this purpose, being based on a variety of function approximation techniques (e.g., \cite{Sutton_1996, MahadevanMaggioni_2005, KonidarisEtAl_2011}).  For VFAs, a state is represented by a feature vector (rather than a table index), where each element reflects the salience of a real-world feature or stimulus in the world.  Such a representation mimics the activation of cortical neurons, which specialize in responding most in the presence of specific stimuli. Also, many deep learning networks express the presence of features in terms of the activation or probability of activation of their hidden nodes.    Depending on the application, a VFA's input representation may vary, but the \emph{output representation} or prediction target is always the same:  reward value.  The reward value scale ranges from reward (positive values) through neutral (zero value) to punishment (negative values).

In our natural environment, some of the stimuli making up the state correlate directly with future rewarding or punishing events or outcomes.  In animal learning or conditioning terms, such stimuli are referred to as ``excitors''.  ``Inhibitors'', on the other hand, are defined as being correlated with the \emph{absence} of future events or outcomes that are otherwise expected.  This distinction, in use since the days of Pavlov \cite{Pavlov_1927}, plays out in real world circumstances.  The shape and red colouring of an apple (excitor) may indicate a positive future reward, only to be canceled by the discovery of a worm's borehole (inhibitor), or the punishment predicted by the scent of a predator (excitor) may be reduced by the perceived distance of its very faint call (inhibitor).  Essentially, the role of an inhibitor is to cancel a certain excitor's predicted outcome and its perceived future reward value (i.e., sending it toward zero).  


Inhibitors and excitors of opposite valence are similar and yet different in certain regards.  A juicy apple (excitor) hanging from a tree beside which lays a predator (excitor of opposite valence) cancels the benefit one might have hoped to receive, just like an inhibitor.   The excitor of opposite valence alone, however, encourages some action (e.g., escape from predator), whereas the inhibitor alone does not because it predicts a neutral outcome.  In a classical conditioning paradigm referred to as conditioned inhibition \cite{Pavlov_1927, Rescorla_1969}, animals receive alternated presentations of a stimulus followed by a reinforcer, say, a reward (A+), and a combination of that stimulus and another, followed by no reward (AX-).  Stimulus A becomes excitatory and stimulus X cancels or inhibits A's reward prediction.  However, the two are not viewed as entirely symmetric opposites in the classical conditioning field.  For example, the extinction of the inhibitor (i.e., eliminating its inhibitory properties) is generally more complicated than mere non-reinforced presentations \cite{ZimmerhartRescorla_1974, LysleFowler_1985}, which is effective on excitatory stimuli such as A.  Thus the extinction of an inhibitor and an excitor of opposite valence are generally not accomplished using the same procedure. 

It is very easy to create confusion between an inhibitor and an excitor of opposite valence in standard VFAs.  We will show that when a VFA is trained on the same data as in the conditioned inhibition paradigm (i.e., [1 0]$\leadsto$1, [1 1]$\leadsto$0), it will subsequently predict a negative reinforcement or punishment when presented with X alone (i.e., [0 1]$\leadsto$-1) instead of a zero-valued outcome, as the conditioned inhibition paradigm suggests.  Indeed, it is possible to have a combination of rewarding and punishing outcomes, which might seem to similarly balance out (i.e., [1 1]$\leadsto$0, where one stimulus is a predictor reward and one is a predictor of punishment).  However, this double outcome is qualitatively very different from the conditioned inhibition case, which captures the omission of an expected outcome.  It would be best if there was a way to distinguish between these cases, instead of always assuming one or the other.  Besides imposing an increase in prediction error, such an assumption could be read incorrectly by an agent and lead to irrational actions (e.g., fear-related responses instead of disappointment-related responses).


Here, we demonstrate this error in concrete terms and offer a simple, generic solution.  Instead of predicting the standard reward value, we train two function approximators: one to predict future punishment and one to predict future reward.  Each prediction is also rectified (i.e., negative values are set to zero).  The typical reward value is finally computed by subtracting the punishment prediction from the reward prediction.  Essentially, instead of allowing positive and negative rewards to cancel each other out and go unnoticed, they are both acknowledged and used to train the model, although the final reward signal used to make decisions does combine them in the usual way.  This allows the model to make the distinction between the omission of one outcome and the prediction of the oppositely valenced outcome.  

Like the standard VFA, separating the prediction of positive and negative reinforcements recognizes the importance of developing high-level representations of the world as an input to reinforcement learning, though doing so is not the focus of the present work.  Instead, we show that the choice of what a learner predicts, or its \emph{output representation}, is equally important, and is reflected in the resulting effect on prediction accuracy.

\section{Predicting rewarding and punishing outcomes separately}
\begin{figure}[h]
\begin{center}
\includegraphics[scale=0.27]{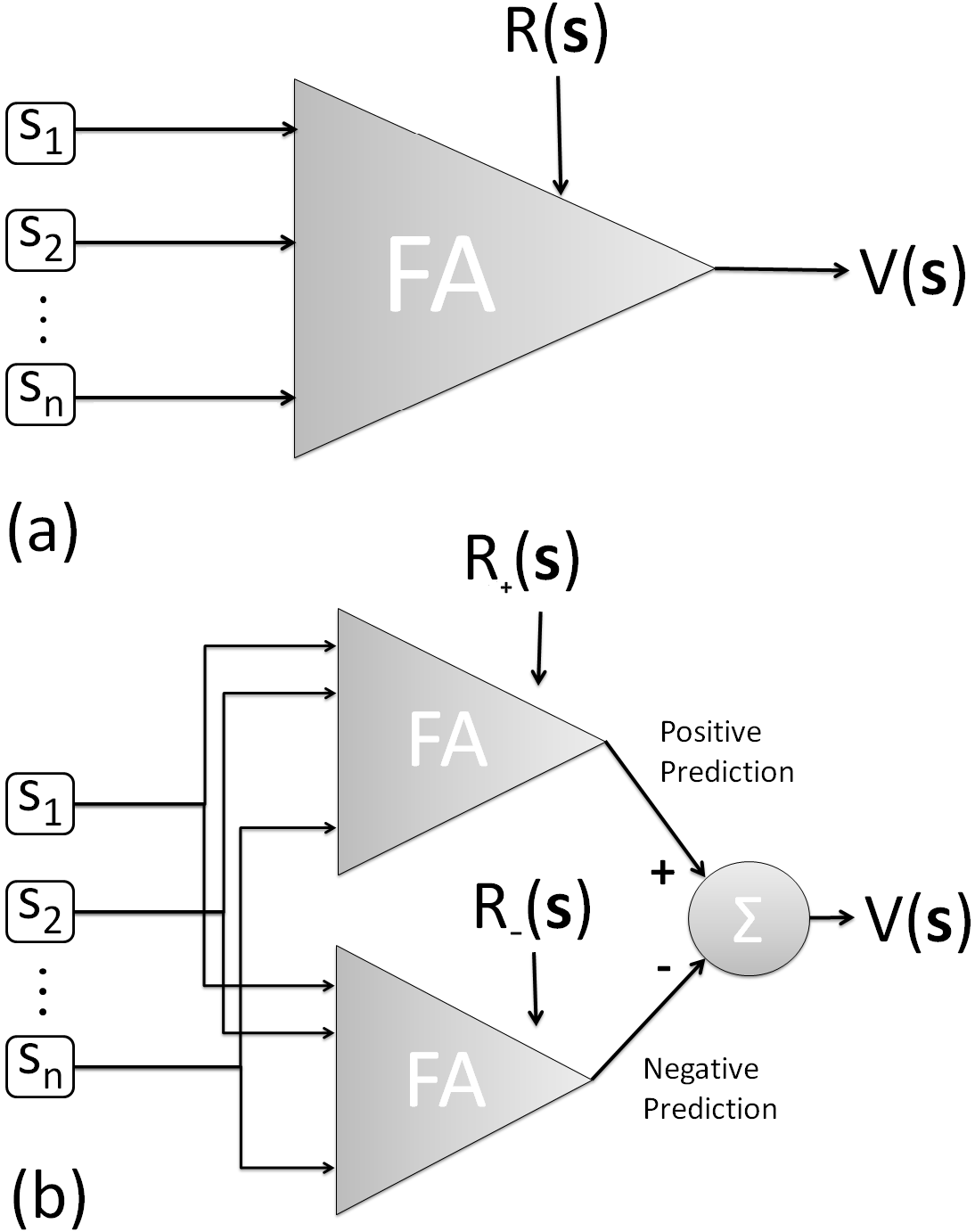}
\end{center}
\caption{\label{fig:TwoStage} VFA architectures for RL using function approximators (FA). Both architectures receive a feature vector input representing the state ($S_1...S_n$), predict future reward value (V(s)), and learn from reinforcements (R(s)).  (a) The typical VFA takes features of the current state and directly predicts the total expected future reward for the agent.  (b) The proposed two-stage VFA architecture: the first stage contains two predictors, one for the expected future rewards and one for the expected future punishments, which are rectified to output positive values only.  In the second stage, the final expected reward value is calculated as the expected future reward minus the expected future punishment.}
\end{figure}

Figure~\ref{fig:TwoStage}A depicts the typical VFA used in RL.  The input features which express the environmental state are submitted to the approximator and it predicts the total expected future reward value from that state.  It is updated according to one of a variety of RL paradigms (e.g., Sarsa, Q-learning, TD$(\lambda)$, etc.) and the specifics of the approximator (e.g., gradient descent for a linear approximator).  In contrast, Figure~\ref{fig:TwoStage}B depicts the deeper, two-staged VFA architecture that this work advocates: two supervised learning models, one that predicts future reward and one that predicts future punishment.  A key difference between these two models' prediction targets and the reward value target of the VFA is that the former has positive values only (a reward or punishment is either expected or not) whereas the latter can have both positive and negative values.  This rectification is a crucial detail of the proposed approach.

This breaking-down of the prediction into sub-predictions is not new to RL \cite{NguyenWidrow_1990, Sutton_1990, Schmidhuber_1990, Schmidhuber_1990b, SchmidhuberHuber_1991, SuttonTanner_2004, RafolsEtAl_2005}.  Although more computation is required in such an approach than for predicting reward alone, the richer predictive output has potential uses, including being able to foresee future states and allow the agent to act in anticipation of them.  In Schmidhuber's work \cite{Schmidhuber_1990b}, a critic (i.e., a VFA) is presented that learns to predict multiple reward/punishment types and the associated prediction errors are backpropagated through a controller module to improve action selection. Applied to a pole-balancing task, the critic's predictive targets reflect different types of pain, or ways in which the task can fail. Schmidhuber found that using these multiple predictive targets allowed the controller to balance the pole with substantially fewer training episodes. In Nguyen and Widrow \cite{NguyenWidrow_1990}, a model is trained to predict the next state values with the ultimate goal of informing a steering controller that will minimize the final error (difference between the ideal final position and orientation of the truck and its location at the end of a series of backup cycles). In Schmidhuber and Huber \cite{SchmidhuberHuber_1991}, the state is represented by pixel intensity sums organized to coarsely emulate a fovea. Here, a model is trained to predict the next state, that is, the pixel sums after a simulated eye movement. Also, in this paper, the model is used to inform a controller and help direct movement effectively.  Neural networks such as those used in these papers to predict the future state could also be used in our proposed approach to predict rewards and punishments separately, albeit with some minor changes. In particular, the pain-predicting critic \cite{Schmidhuber_1990b} learns to predict reward-related outcomes, as does our architecture, except that instead of predicting many forms of one kind of valence, we subdivide the problem into the prediction of rewards and the prediction of punishments.  This will allow us to prevent confusion between excitors and inhibitors of opposite valence.


\section{Function approximators}
In the evaluations that follow, we consider several forms of learning models to show generally the effect of the standard VFA and two-stage VFA architectures in specific scenarios.  The VFA models evaluated are based on the following four supervised learning approaches: least mean squares (LMS), support-vector regression (SVR) \cite{SmolaScholkopf_2004}, classification and regression trees (CART), and multi-layer perceptron.  All of these, except LMS, are capable of representing non-linear functions.   With the simple task evaluated, these models required very little adjusting of parameters, which might otherwise complicate the interpretation of the results.  For the two approximators in the two-stage architecture, we employed a rectified version of LMS (rLMS, derived below), LMS combined with a logistic/sigmoid transfer function, and Dual Pathway Regression (DPR) \cite{ConnorTrappenberg_2013}.  All three models, in one way or another, rectify their outputs.

\subsection{Standard VFA function approximators}
The form of SVR used here, called epsilon-SVR, involves fitting a hyperplane with margins (i.e., a hyper-rectangular prism) so that it encompasses all data while at the same time minimizing the hyperplane's slope \cite{SmolaScholkopf_2004}.  This is often infeasible, so that outliers are permitted but at a cost, leading to a trade-off between minimization of the slope and the acceptance of outliers.  Since explanations of simulation results will not rely on the formal definition of this non-linear model, we direct the interested reader to Smola and Sch\"olkopf \cite{SmolaScholkopf_2004}.  We used LIBSVM \cite{ChangLin_2011} in our simulations (Settings: $cost=10$, $epsilon=1e-5$, radial basis function kernel with $\gamma=1/No.\,Features$).  Although SVR will serve as the primary and most effective among the VFAs, we also tested LMS, CART \cite{BreimanEtAl_1984} (MATLAB Statistics toolbox) and MLP \cite{RumelhartEtAl_1986} (MATLAB Neural Networks toolbox) to show the generality of our results.

\subsection{Two-stage VFA architecture function approximators}
We will now derive the rLMS regression model used in the two-stage architecture by beginning with LMS itself.  Underlying LMS linear regression is the goal to maximize the likelihood of the given data being generated by a Gaussian random variable with a parameterized n-dimensional linear mean.  This function can be expressed as  
\begin{equation}
\label{eq:linfun}
y = \phi^Tx + \mathcal{N}(0, \sigma)
\end{equation}
where $y$ is the outcome being predicted, $x$ is a vector of inputs, $\phi$ is the vector of model parameters, and 0 and $\sigma$ represent the mean and standard deviation of the Gaussian random variable, respectively.  The probability density function (PDF) becomes
\begin{equation}
\label{eq:GRVLMPDF}
p(y, x|\phi) = \frac{1}{\sqrt{2\pi}\sigma}e^{-\frac{(y - \phi^Tx)^2}{2\sigma^2}}
\end{equation}
For the two-stage architecture, the reinforcer outcome being predicted cannot be less than zero, 
\begin{equation}
\label{eq:linfun2}
y = G(\phi^Tx + \mathcal{N}(0, \sigma))
\end{equation}
where $G()$ is the threshold-linear function \cite{NairHinton_2010} which returns its argument whenever it is positive and returns zero otherwise.  The probability function now becomes
\begin{equation}
\label{eq:GRVLMPDF2}
p(y, x|\phi) = \begin{cases}
\begin{array}{cc}
\frac{1}{\sqrt{2\pi}\sigma}e^{-\frac{(y - \phi^Tx)^2}{2\sigma^2}}, & for\:y>0\\
\frac{1}{2}(1-{\rm erf}(\frac{\phi^Tx}{\sqrt{2}\sigma}), & for\:y=0\\
0, & for\:y<0
\end{array}\end{cases}
\end{equation}
differing from the PDF in Equation~\ref{eq:GRVLMPDF} by zeroing the probability of negative $y$ values and instead heaping it onto the $y=0$ case.  That is, this probability function is a density over $y>0$ plus a point mass at $y=0$, which is equal to 1 minus the probability of $y>0$.  From this function, let us derive the rLMS learning rule and, indirectly, the LMS learning rule as well.  Given data generated from Equation~\ref{eq:linfun2}, we can infer the values of $\phi$ using maximum likelihood estimation.  The probability or likelihood that a certain training data set is generated from the probability function in Equation~\ref{eq:GRVLMPDF2} is
\begin{eqnarray}
\label{eq:rLMSlh}
L(\phi) = p(y^{(1)}, ...y^{(m)}, x^{(1)}, ...x^{(m)}| \phi) \nonumber \\ = \prod_{i=1}^m p(y^{(i)}, x^{(i)}|\phi)
\end{eqnarray}
where $m$ is the number of training data points.  We can maximize this convex likelihood function by taking its log and ascending its gradient,
\begin{eqnarray}
\log\, L(\phi) & = \log \prod_{i=1}^m p(y^{(i)}, x^{(i)}|\phi) = \sum_{i=1}^m \log\, p(y^{(i)},x^{(i)}|\phi)\nonumber \\
	      & = -m_{y>0}\log(\sqrt{2\pi} \sigma) - \frac{1}{2\sigma^2}\sum_{i, y^{(i)}>0} (y^{(i)}- \phi^Tx^{(i)})^2 \nonumber \\
	      & - m_{y=0}\log(2) + \sum_{i, y^{(i)}=0}\log(1-\operatorname{erf}(\frac{\phi^Tx^{(i)}}{\sqrt{2}\sigma}))
\end{eqnarray}
where $m_{y>0}$ and $m_{y=0}$ are the number of data points when $y>0$ and $y=0$, respectively.  Taking the gradient of this function with respect to each $\phi_j$ gives
\begin{equation}
\label{eq:derlog}
\frac{\partial\, \log\, L(\phi)}{\partial\phi_{j}}= 
\frac{1}{\sigma^2}\sum_{i, y^{(i)}>0}(y^{(i)}-\phi^{T}x^{(i)})x_{j}^{(i)} - \sqrt{\frac{2}{\pi\sigma^2}}\sum_{i, y^{(i)}=0}\frac{e^{-\frac{(\phi^Tx^{(i)})^2}{2\sigma^2}}}{1-\operatorname{erf}(\frac{\phi^Tx^{(i)}}{\sqrt{2}\sigma})}x_{j}^{(i)}
\end{equation}
The gradient can be ascended by iteratively updating the model parameters,
\begin{equation}
\label{eq:gradup}
\phi_j =: \phi_j + \alpha \frac{\partial\, \log\, L(\phi)}{\partial \phi_j}
\end{equation}
where the learning rate, $\alpha = \frac{\sigma^2}{n+2}$.  The learning rule can be similarly derived for standard LMS (Equation~\ref{eq:GRVLMPDF}), resulting in
\begin{equation}
\label{eq:derlog2}
\frac{\partial\, \log\, L(\phi)}{\partial\phi_{j}}= 
\frac{1}{\sigma^2}\sum^M_{i=1}(y^{(i)}-\phi^{T}x^{(i)})x_{j}^{(i)}
\end{equation}
where $M$ is the total number of data points.  The  practical feature of the rectified learning rule that distinguishes it is what happens for data points where $y=0$:  when the weighted sum $\phi^Tx$ is negative, learning is essentially deactivated, especially for small $\sigma$.  For our simulations, we use a low noise variance of $\sigma=1e-4$.

The other two function approximators used in the two-stage architecture, DPR and LMS with a sigmoid transfer function, can be similarly derived.  All that is different are the PDFs that the models embody.  In the case of LMS with a sigmoid transfer function, 
\begin{equation}
\label{eq:LMSsig}
p(y, x|\phi) =\frac{1}{\sqrt{2\pi}\sigma}e^{-\frac{(y - \frac{1}{1 + e^{-\phi^Tx}}))^2}{2\sigma^2}}
\end{equation}
and for DPR \cite{ConnorTrappenberg_2013},
\begin{equation}
\label{eq:DPR}
p(y, x|\phi) = \frac{1}{\sqrt{2\pi}\sigma}e^{-\frac{(y - 2|\phi^T_+|x(1 - \frac{1}{1 + e^{-|\phi^T_-|x}}))^2}{2\sigma^2}}
\end{equation}
where $\phi^T_+$ and $\phi^T_-$ are separate sets of parameters to be inferred, representing the positive and negative pathways of the dual pathway approach, respectively.

%
%

\section{Simulation and results}
To show the difference between the two architectures, a single-step RL problem is simulated.   It is the combination of two simultaneous conditioned inhibition experiments, where there are both rewarding and punishing excitatory stimuli/reinforcements as well as inhibitory stimuli.  The experiment is arranged to show how well these architectures distinguish between inhibitory features and excitatory features of the opposite valence.  As a RL problem, there are no actions to take; an agent is forced from one state to the next.  As an aside, however, one can imagine an agent's actions also being represented as input features, for which the process of learning to predict outcomes based on proposed actions would then be the same.

The input data is comprised of 50 samples with four features apiece: two excitors, one for reward ($E^{+}$) and one for punishment ($E^{-}$) and two inhibitors, one canceling reward ($I^{+}$) and one canceling punishment ($I^{-}$) which are each given real-values between 0 and 1. Each data point can be classified as either ``linear'' where the inhibitory feature saliences do not exceed their associated excitatory feature saliences ($E^{+} \geq I^{+}$ and $E^{-} \geq I^{-}$) or ``non-linear'', otherwise.  The expected future reward (or reward value) output is computed as R(s) = G($E^{+} - I^{+}) - G(E^{-} - I^{-})$, where $G()$ function again is a rectifier, so that an inhibitory feature alone cannot have a negative value.

The learning cycle for the standard VFA in this experiment is the following:
\begin{itemize}
\item the 4-feature state vector is presented as input.
\item the reward value prediction is made 
\item in the next time step, the actual reward value outcome ($R(s)$) is received and used to update the model
\end{itemize}

The learning cycle for the two-stage VFA architecture in this experiment is the following:
\begin{itemize}
\item the 4-feature state vector is presented as input.
\item the prediction of the positive reinforcement is made by one function approximator and the prediction of the negative reinforcement is made by a second approximator
\item the reward value prediction is computed as the positive reinforcer prediction minus the negative reinforcer prediction 
\item in the next time step, the absence/presence of each reinforcer is detected and used to update the associated function approximator.
\end{itemize}

\begin{figure}[h]
\begin{center}
\includegraphics[scale=0.33]{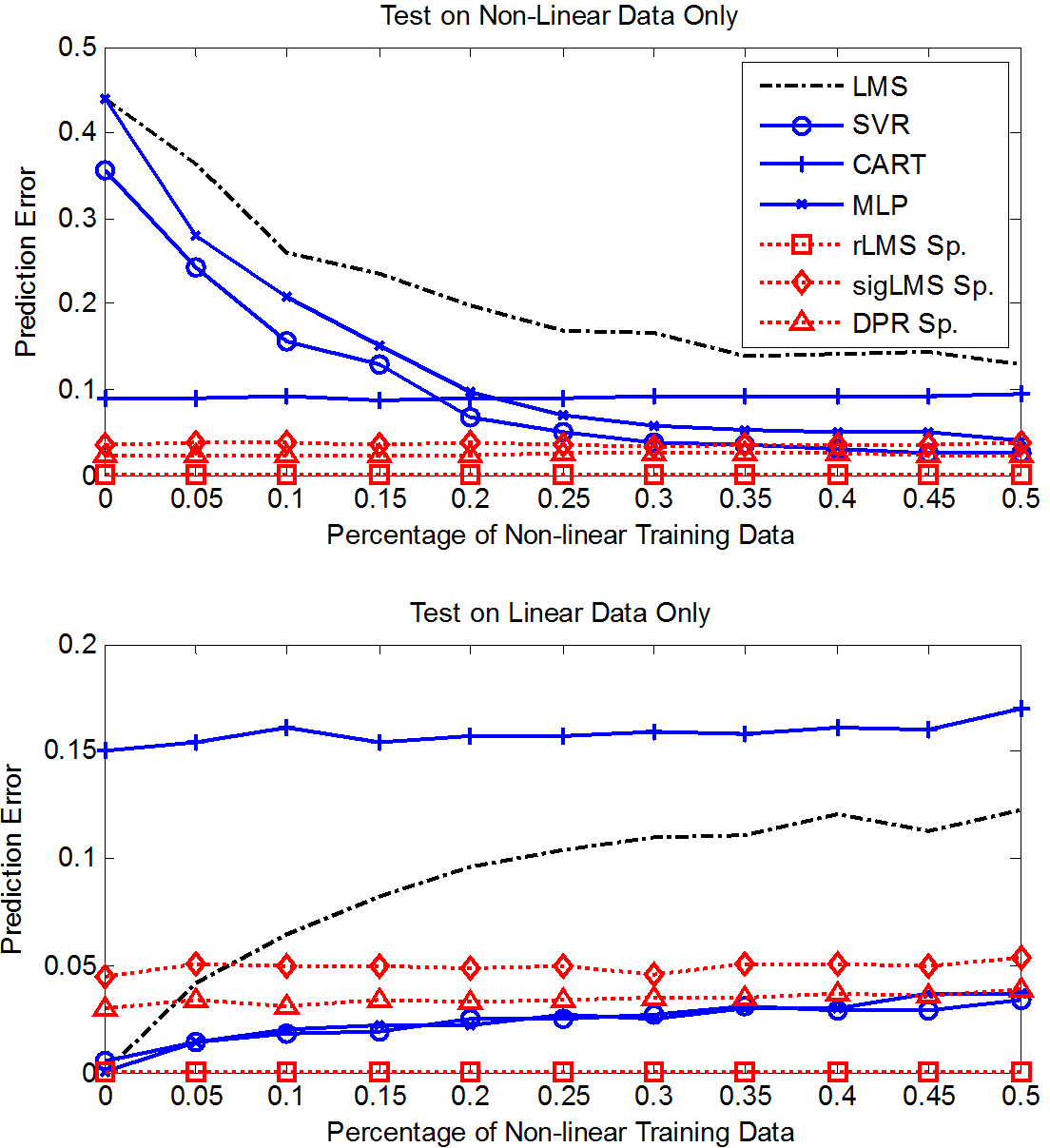}
\end{center}
\caption{\label{fig:ResultsGeneral} Testing standard VFAs and the two-stage VFA architecture in the dual-valenced conditioned inhibition simulation.  In the upper panel, we see the two-stage results (red) perform well with low numbers of non-linear training data when the test data is non-linear relative to the VFA results (blue).  The sigmoid and DPR functions do not perform as well as rLMS, however, particularly when the test data is linear (lower panel) because these models have inherent non-linearities that affect them in this scenario.}
\end{figure}

Figure~\ref{fig:ResultsGeneral} shows the mean prediction error as the percentage of non-linear training samples increases among the 50 data points used, showing results for where the \emph{test data} is either completely ``linear'' (lower panel) or completely ``non-linear'' (upper panel).  The simulation was run 20 times for each 5\% increment.  Standard errors are left out for clarity's sake.  With low amounts of non-linear training data, non-linear test results suffer significantly for all of the VFAs (blue), whereas the two-stage architecture approaches (red) fare much better, especially rLMS, which is directly modeled to solve this problem.  The two-stage architecture function approximators other than rLMS do not perform as well when the test data is ``linear'' because the logistic and DPR functions have inherent non-linearities that prevent them from reaching zero prediction error.  The CART result is better than the other VFAs when non-linear training data is slight.  Its constant level of effectiveness may be due to the fact that its ability to make sharply defined rules allows it to roughly align with the contours of the present world model.  Nevertheless, it does not achieve as good results as do the two-stage architecture models.

\begin{figure}[h]
\begin{center}
\includegraphics[scale=0.32]{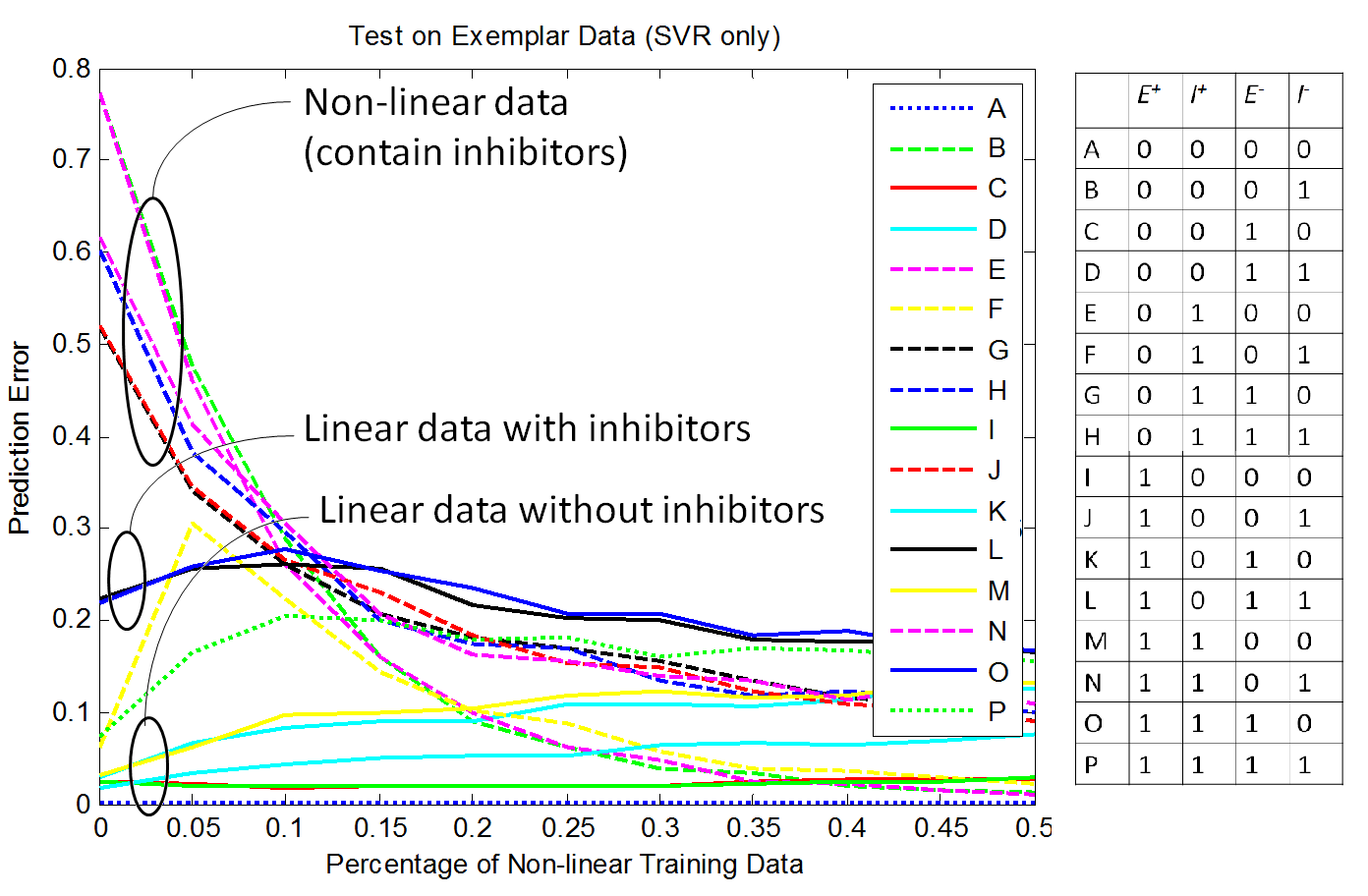}
\end{center}
\caption{\label{fig:ResultsSpecific} A breakdown of SVR (i.e., standard VFA) prediction error by canonical data points shown in the table next to the graph.  SVR has trouble properly predicting the state-value of ``non-linear'' data points when there are few such points in the training data.  Even ``linear'' data points that include inhibitors have higher prediction errors than those that do not.  Contrast this with the two-stage VFA architecture employing rLMS, where there is essentially zero error for every canonical data point, for every percentage of non-linear training data.}
\end{figure}

Figure~\ref{fig:ResultsSpecific} provides a more detailed picture of where the VFA errors arise.  In the table to the right, there are 16 canonical data points representing all general cases in which data fall, which are tested after training with different levels of ``non-linear'' data, as before.  The largest prediction errors occur in SVR for non-linear data points (i.e., where inhibitors play a role, long-dashed results) when there are few non-linear training data points.  The next highest prediction errors occur for data points that are linear (solid lines) but contain inhibitors as well (e.g., ``O'', which contains a positive inhibitor to cancel the prediction of a reward).  Finally, linear data in the absence of inhibitors fare best.  As expected, VFAs see the inhibitor as an excitor of opposite valence rather than as a stimulus that cancels an expected outcome.  Take canonical data point `B', for example.  It has a single inhibitor activated (indicating the absence of a punishment) and no excitors, which ought to predict zero value.  However, with low levels of non-linear data, it is seen to have substantial positive reward value.  So, when trained with linear data, even a non-linear VFA will see the inhibitor as a linear excitatory feature of the opposite valence in order to ``cancel out'' its truly excitatory companion.

\section{Discussion}
As the results show, the presence of inhibitors, without much non-linear training data, causes confusion in the standard VFA.  Even with half the training data being non-linear, prediction errors are still generally above those when using a two-stage architecture, which maintain relatively low prediction errors regardless of the amount of non-linear data.  With little non-linear data, the VFA sees the inhibitor in a linear sense, as an excitatory stimulus with opposite valence.  Why does the two-stage architecture work?  Recall that each function approximator predicts the presence/absence of a single reinforcer and that the prediction cannot be negative (e.g., less than zero food) because the output is rectified.  Without this rectification (i.e, using standard LMS), the two-stage architecture would give similarly poor results.  In short, inhibitors are not permitted to make negative predictions (via rectification), but only cancel their associated prediction.  With enough training data or experience, we see that the standard VFAs are capable of learning the relationship between excitors and inhibitors.  The two-stage VFA architecture, however, essentially encodes the potential of such relationships and avoids the need to explicitly learn them.  This results in correct perception of the reinforcement landscape sooner, which may lead to better early decisions. 

It may be that certain reinforcement learning agents or robots will rarely meet with inhibitors and thus not be given opportunity to experience the confusion demonstrated.  Yet, it still may be appropriate to include a two-stage VFA architecture, since doing so does little to disturb the general case.  When using linear training data only, prediction errors were still low for the two-stage architecture, particularly for the use of rLMS.  The two-stage VFA architecture, then, insures the agent against confusion in the presence of inhibitors.


One feature of the function approximators used with the two-stage architecture is that they essentially shut off learning when presented with negative predictors (i.e., inhibitors) alone and thereby not change their inhibitory qualities.  This resonates with the animal learning literature (but see \cite{BaetuBaker_2010}), which generally finds that non-reinforced presentations of such inhibitors do not extinguish them (i.e., they keep their inhibitory quality).  This is in contrast to the presentation of excitors alone, which cause them to lose their excitatory quality and to become ``neutral'', something that the two-stage function approximators also emulate.



It is possible that the first stage of the two-stage architecture could contain multiple function approximators to predict as many types of rewards or punishments as desired, much like Schmidhuber's \cite{Schmidhuber_1990b} predictor of varying forms of pain.  One reason for grouping multiple forms of pain together as one, however, is that it appears that this is what biological systems may do.  Bakal et al. \cite{BakalEtAl_1974} found in a classical conditioning blocking experiment that subjects did not associate predictors with particular punishments but rather a common negative outcome.

Apart from the rectification, the function approximators we used to demonstrate the two-stage VFA architecture learn primarily linear relationships.  However, this is not to say that one could not use in their place models that learn non-linear relationships.  In fact, it will be necessary to have non-linear predictive abilities for specific combinations of features (e.g., XOR or AND cases).  Importantly, non-linear function approximators that are used in the two-stage architecture must also rectify their output to enjoy its benefits.  However, such rectification is non-native to most function approximators and would have to be carefully incorporated.  

The input representation that would best support the two-stage architecture is likely a hierarchical one.  The best deep learning representations would seem to be those with varying-levels of feature complexity, with specific features that correlate with either rewards or punishments meaningful to the agent.  For the thirsty agent, a low-level bank of edge-detectors would seem sufficient to identify a pool or puddle with numerous ripples.  In the case of a calm pool, however, a very high-level representation that captures the mirror-like effect of the water would be more discriminative.  A hierarchical representation affords such varying levels of complexity and thus would seem to effectively support the prediction of future reinforcements or omission thereof.

\section{Conclusion}
We have demonstrated how a two-stage VFA architecture can overcome a fundamental issue faced by standard VFAs: the inability to distinguish between predictors of reinforcement and inhibitors of reinforcement of the opposite valence with little training data.  The prediction of reward value is replaced by the prediction of rewards as a group and punishments as a group, which themselves have a positive salience only.  Such a rectified output representation allows an inhibitor (or predictor of omission) to have a negative influence when presented together with an excitor of the same valence, but no influence when it is presented alone.  This eliminates its confusion with predictors of reinforcement with the opposite valence.  When used with the right function approximator, the two-stage VFA architecture ensures the reinforcement learning agent appropriately handles inhibitors without increasing prediction errors much in the general case (i.e., no inhibitors).  This suggests that RL agents may stand to benefit from incorporating the two-stage VFA architecture even if this situation is unlikely, just in case they run into such a situation.

\section*{Acknowledgments}
This work was supported by funding from NSERC and Dalhousie University.

\section*{References} 
\bibliography{ICLRPaper}

%

\end{document}